\documentstyle[graphicx,times,mathptm]{IJCNNarticle}

\newcommand\be{\vspace*{-1pt}\begin{equation}}
\newcommand\ee{\end{equation}\vspace*{-1pt}}
\newcommand\knn{$k$-NN}

\def\bO{{\mathbf O}} \def\bR{{\mathbf R}}  
\def\bX{{\mathbf X}}  \def\bY{{\mathbf Y}}

\def\knn{{$k$-NN}} 
\def\R{{\cal R}}   \def\S{{\cal S}}  \def\T{{\cal T}} \def\V{{\cal V}}

\begin{document}

\title{Meta-learning: searching in the model space.}

\author{W{\l}odzis{\l}aw Duch and Karol Grudzi\'nski\\
Department of Computer Methods, Nicholas Copernicus University, \\
Grudzi\c{a}dzka 5, 87-100 Toru\'n, Poland. \\
WWW: http://www.phys.uni.torun.pl/kmk }

\maketitle
\thispagestyle{empty}\pagestyle{empty}

\begin{abstract}
There is no free lunch, no single learning algorithm that will outperform other algorithms on all data. In practice different approaches are tried and the best algorithm selected. An alternative solution is to build new algorithms on demand by creating a framework that accommodates many algorithms. The best combination of parameters and procedures is searched here in the space of all possible models belonging to the framework of Similarity-Based Methods (SBMs). Such meta-learning approach gives a chance to find the best method in all cases. Issues related to the meta-learning and first tests of this approach are presented.
\end{abstract}

\section{Introduction.}

The `no free lunch' theorem \cite{Duda} states that there is no single learning algorithm that is inherently superior to all the others. The back side of this theorem is that there are always some data on which an algorithm that is evaluated give superior results. Neural and other computational intelligence methods are usually tried on a few selected datasets on which they work well. A review of many approaches to classification and comparison of performance of 20 methods on 20 real world datasets has been done within the {\sf StatLog} European Community project \cite{statlog}. The accuracy of 24 neural-based, pattern recognition and statistical classification systems has been compared on 11 large datasets by Rohwer and Morciniec \cite{Rohwer}. No consistent trends have been observed in the results of these large-scale studies. Frequently simple methods, such as the nearest neighbor methods or $n$-tuple methods, outperform more sophisticated approaches \cite{Rohwer}.

In real world applications a good strategy is to find the best algorithm that works for a given data trying many different approaches. This may not be easy. First, not all algorithms are easily available, for example there is no research or commercial software for some of the best algorithms used in the {\sf StatLog} project \cite{statlog}. Second, each program requires usually a different data format. Third, programs have many parameters and it is not easy to master them all. Our strategy is to use a framework for Similarity-Based Methods (SBM) introduced recently \cite{framework,applied}.
The meta-learning approach described here involves a search for the best model in the space of all models that may be generated within SBM framework. Simplest model are created at the beginning and new types of parameters and procedures are added, allowing to explore more complex models. Constructive neural networks as well as genetic procedures for selection of neural architectures increase model complexity by adding the same type of parameters, generating different models within a single method. Meta-learning requires creation of models based on different methods, introducing new types of parameters and procedures.

In the next section we will briefly introduce the SBM framework, explain which methods may be generated within SBM, present the meta-learning approach used to select the best method and describe our preliminary experiences analyzing a few datasets. Conclusions and plans for future developments close this paper.  Although the meta-learning approach is quite general this paper is focused on classification methods.

\section{A framework for meta-learning} \label{NN-MD}

{\bf An algorithm}, or {\bf a method}, is a certain well-defined computational procedure, for example an MLP neural network or \knn\ method. {\bf A model} is an instance of a method with specific values of parameters.
A framework for meta-learning should allow for generation and testing of models derived from different methods. It should be sufficiently rich to accommodate standard methods.
The SBM framework introduced recently \cite{framework,applied} seems to be most suitable for the purpose of meta-learning. It covers all methods based on computing similarity between the new case and cases in the training library.  It includes such well-known methods as the $k$--Nearest Neighbor (\knn) algorithm and it's extensions, originating mainly from machine learning and pattern recognition fields, as well as neural methods such as the popular multilayer perceptron networks (MLP) and networks based on radial--basis functions (RBF).

Methods that belong to the SBM are based on specific parameterization of the $p(C_i|\bX;M)$ posterior classification probability, where the model $M$ involves various procedures, parameters and optimization methods. Instead of focusing on improving a single method a search for the best method belonging to the SBM framework should select optimal combination of parameters and procedures for a given problem.

Below $N$ is the number of features, $K$ is the number of classes, vectors are in bold faces while vector components are in italics. The following steps may be distinguished in the supervised classification problem based on similarity estimations:\\
1) Given a set of objects (cases) $\{\bO^p\}$, $p=1..n$ and their symbolic labels $C(\bO^p)$, define useful numerical features $X_j^p=X_j(\bO^p), j=1...N$ characterizing these objects. This preprocessing step involves computing various characteristics of images, spatio-temporal patterns, replacing symbolic features by numerical values etc.\\
2) Find a measure suitable for evaluation of similarity or dissimilarity of objects represented by vectors in the feature space, $D(\bX,\bY)$.  \\
3) Create a reference (or prototype) vectors $\bR$ in the feature space using the similarity measure and the training set $\T=\{\bX^p\}$ (a subset of all cases given for classification). \\
4) Define a function or a procedure to estimate the probability $p(C_i|\bX;M), i=1..K$ of assigning vector $\bX$ to class $C_i$. The set of reference vectors, similarity measure, the feature space and procedures employed to compute  probability define the classification model $M$.\\
5) Define a cost function $E[\T;M]$ measuring the performance accuracy of the system on a training set $\T$ of vectors; a validation set $\V$ composed of cases that are not used directly to optimize model $M$ may also be defined and performance $E[\V;M]$ measuring generalization abilities of the model assessed.\\
6) Optimize the model $M_a$ until the cost function $E[\T;M_a]$ reaches minimum on the set $\T$ or on the validation set  $E[\V;M_a]$. \\
7) If the model produced so far is not sufficiently accurate add new procedures/parameters creating more complex model $M_{a+1}$.\\
8) If a single model is not sufficient create several local models $M_a^{(l)}$ and use an interpolation procedure to select the best model or combine results creating ensembles of models.

All these steps are mutually dependent and involve many choices described below in some details. The final classification model $M$ is build by selecting a combination of all available elements and procedures.
A general similarity-based classification model may include all or some of the following elements:\\[3pt]
$M=\{\bX(\bO), \Delta(\cdot,\cdot), D(\cdot,\cdot), k, G(D), \{\bR\}, \{p_i(R)\}, E[\cdot]$, $K(\cdot), \S(\cdot)\}$, where: \\
$\bX(\bO)$ is the mapping defining the feature space and selecting the relevant features;\\
$\Delta_j(X_j;Y_j)$ calculates similarity of $X_j,\ Y_j$ features, $j=1 .. N$; \\
$D(\bX,\bY)=D(\{\Delta_j(X_j;Y_j)\})$ is a function that combines similarities of features to compute similarities of vectors; if the similarity function selected has metric properties the SBM may be called the minimal distance (MD) method. \\
$k$ is the number of reference vectors taken into account in the neighborhood of $\bX$; \\
$G(D)=G(D(\bX,\bR))$ is the weighting function estimating contribution of the reference vector $\bR$ to the classification probability of $\bX$;\\
$\{\bR\}$ is a set of reference vectors created from the set of training vectors $\T=\{\bX^p\}$ by some selection and optimization procedure; \\
$p_i(\bR), i=1..K$ is a set of class probabilities for each reference vector;\\
$E[\T;M]$ or $E[\V;M]$ is a total cost function that is minimized at the training stage; it may include a misclassification risk matrix $\R(C_i,C_j), i,j =1..K$;\\
$K(\cdot)$ is a kernel function, scaling the influence of the error, for a given training example, on the total cost function;\\
$\S(\cdot)$ is a function (or a matrix) evaluating similarity (or more frequently dissimilarity) of the classes; if class labels are soft or are if they are given by a vector of probabilities $p_i(\bX)$ classification task is in fact a mapping.
$\S(C_i,C_j)$ function allows to include a large number of classes, ``softening" the labeling of objects that are given for classification.

Various choices of parameters and procedures in the context of network computations leads to a large number of similarity-based classification methods. Some of these models are well known and some have not yet been used.
We have explored so far only a few aspects of this framework, describing various procedures of feature selection, parameterization of similarity functions for objects and single features, selection and weighting of reference vectors, creation of ensembles of models and estimation of classification probability using ensembles, definitions of cost functions, choice of optimization methods, and various network realizations of the methods that may be created by combination of all these procedures \cite{applied,SBMCC,ustron99,ijcnn99}. A few methods that may be generated within the SBM framework are mentioned below.

The \knn\ model $p(C_i|\bX;M)$ is parameterized by $p(C_i|\bX;k,D(\cdot),\{\bX\}\})$, i.e. the whole training dataset is used as the reference set, $k$ nearest prototypes are included with the same weight, and a typical distance function, such as the Euclidean or the Manhattan distance, is used.  Probabilities are $p(C_i|\bX;M)=N_i/k$, where $N_i$ is the number of neighboring vectors belonging to the class $C_i$. The most probable class is selected as the winner. Many variants of this basic model may be created.  Instead of enforcing exactly $k$ neighbors the radius $r$ may be used as an adaptive parameter. Using several radial parameters and the hard-sphere weighting functions Restricted Coulomb Energy (RCE) algorithm is obtained \cite{RCE}. Selection of the prototypes and optimization of their position leads to old and new variants of the LVQ algorithms \cite{Kohonen}. Gaussian classifiers, fuzzy systems and RBF networks are the result of soft-weighting and optimization of the reference vectors. Neural-like network realizations of the RBF and MLP types are also special cases of this framework \cite{SBMCC}.

The SBM framework is too rich that instead of exploring all the methods that may be generated within it an automatic search for the best method and model is needed: a meta-learning level.

\section{Meta-learning issues} \label{ML-issues}

Parameters of each model are optimized and a search is made in the space of all models $M_a$ for the simplest and most accurate model that accounts for the data. Optimization should be done using validation sets (for example in crossvalidation tests) to improve generalization. Starting from the simplest model, such as the nearest neighbor model, qualitatively new ``optimization channel" is opened  by adding the most promising new extension, a set of parameters or a procedure that leads to greatest improvements. Once the new model is established and optimized all extensions of the model are created and tested and another, better model is selected. The model may be more or less complex than the previous one (for example, feature selection or selection of reference vectors may simplify the model). The search in the space of all SBM models is stopped when no significant improvements are achieved by new extensions.

In the case of the standard \knn, the classifier is used with different values of $k$ on a training partition using leave-one-out algorithm and applied to the test partition. The predicted class is computed on the majority basis. To increase the classification accuracy one may  first optimize $k, (k_1 \leq k \leq k_2)$ and select $m \leq k_2 - k_1$ best classifiers for an ensemble model.
In the case of weighted \knn\ either $k$ is optimized first and then best models created optimizing all weights, or best models are selected after optimization for a number of $k$ values (a more accurate, but costly procedure).

Selecting a subset of best models that should be included in an ensemble is not an easy task since the number of
possibilities grows combinatorially and obviously not all subsets may be checked. A variant of the best-first search (BFS) algorithm has been used for this selection.
We have already used the BFS technique for the optimization of weights and for selection of the best attributes \cite{ustron99,ijcnn99}. BFS algorithm can be used for majority voting of models derived from weighted-NN method based on minimization, or based on standard \knn\ with different $k$, or for selection of an optimal sequence of any models.

The evaluation function $C(M_l)$ returns the classification accuracy of the model $M_l$ on a validation set; this accuracy refers to a single model or to an ensemble of models selected so far. Let $N$ denote the initial number of models from which selection is made and $K$ the number of models that should be selected. The model sequence selection algorithm proceeds as follows:

\begin{enumerate}
\item Initialize:

\begin{enumerate}
\item Create a pool of $M$ initial models, ${\cal M}=\{M_l\}, l=1\dots M$.
\item Evaluate all initial models on the validation set, arrange them in a decreasing order of accuracy $C_a(M_i)\geq C_a(M_j)$ for $i>j$.
\item Select the best model $M_1$ from the ${\cal M}$ pool as the reference.
\item Remove it from the pool of models.
\end{enumerate}

\item Repeat until the pool of models is empty:

\begin{enumerate}
\item For each model $M_l$ in the pool evaluate its performance starting from the current reference model.
\item Select the reference + $M_l$ model with highest performance and use it as the current reference; if several models have similar performance select the one with lowest complexity.
\item If there is no significant improvement stop and return the current reference model; otherwise accept the current reference model + $M_l$ as the new reference.
\item Remove the $M_l$ model from the pool of available models.
\end{enumerate}

\end{enumerate}

At each step at most $M - L$ sequences consisting of $L=M-1 .. 1$ models are evaluated. Frequently the gain in performance may not justify additional complexity of adding a new model to the final sequence. The result of this algorithm is a sequence of models of increasing complexity, without re-optimization of previously created models. This ``best-first" algorithm finds a sequence of models that give the highest classification accuracy on validation partition. In case of \knn-like models, calculations may be done on the training partition in the leave-one-out mode instead of the validation partition.

The algorithm described above is prone to local minima, as any ``best-first" or gradient-based algorithm. The beam search algorithm for selection of the best sequence of models is more computationally expensive but it has a better chance to find a good sequence of models. Since the SBM scheme allows to add many parameters and procedures, new models may also be created on demand if adding models created so far does not improve results. Some model optimizations, such as the minimization of the weights of features in the distance function, may be relatively expensive. Re-optimization of models in the pool may be desirable but it would  increase the computational costs significantly.  Therefore we will investigate only the simplest ``best-first" sequence selection algorithm, as described above.

\section{Numerical experiments}

We have performed preliminary numerical tests on several datasets. The models taken into account include optimization of $k$, optimzation of distance function, feature selection and optimization of the weights scaling distance function:
\be
D(\bX,\bY)^\alpha=\sum_{i=1}^n s_i |X_i-Y_i|^\alpha
\ee
At present only $\alpha =1, 2$ is considered (Euclidean and Manhattan weighted functions), and two other distance function, Chebyschev and Camberra \cite{SBMCC}, but full optimization of $\alpha$ should soon be added. Various methods of optimization may be used but we have implemented only the simplex method which may lead to the weighted models with relatively large variance. The goal of further search for the best model should therefore include not only accuracy but also reduction of variance, i.e. stabilization of the classifier.

\subsection{Monk problems}

The artificial dataset Monk-1 \cite{monks} is designed for rule-based symbolic machine learning algorithms (the data was taken from the UCI repository \cite{UCI}). The nearest neighbor algorithms usually do not work well in such cases. 6 symbolic features are given as input, 124 cases are given for training and 432 cases for testing. We are interested here in the performance of the model selection procedures.

The meta-learning algorithm starts from the reference model, a standard \knn, with $k=1$ and Euclidean function. The leave-one-out training accuracy is 76.6\% (on test 85.9\%). At the first level the choice is: optimization of $k$, optimization of the type of similarity function, selection of features and weighting of features. Results are summarized in the Table below. Feature weighting (1, 1, 0.1, 0, 0.9, 0), implemented here using a search procedure with 0.1 quantization step, already at the first level of search for the best extension of the reference model achieves 100\% accuracy on the test set and 99.2\%, or just a single error, in the leave-one-out estimations on the training set. Additional complexity may not justify further search. Selection of the optimal distance for the weighted \knn\ reference model achieves 100\% on both training and the test set, therefore the search procedure is stopped.

\begin{table}
\begin{center}
\caption{Results for the Monk-1 problem with \knn\ as reference model.}

\begin{tabular}{|l|l|l|}
\hline
Method & Acc. Train \% & Test \% \\
\hline\hline
ref = \knn, $k$=1, Euclidean		& 76.6  	& 85.9 \\
ref + $k$=3					& 82.3 	& 80.6 \\
ref + Camberra distance			& 79.8 	& 88.4 \\
ref + feature selection 1, 2, 5		& 96.8 	& 100.0 \\
ref + feature weights			& 99.2 	& 100.0 \\
\hline
ref = \knn, Euclid, weights		& 99.2  	& 100.0 \\
ref + Camberra distance			& 100.0  	& 100.0 \\

\hline
\end{tabular}
\end{center}
\end{table}

In the Monk 2 problem the best combination sequence of models was \knn\ with Camberra distance function, giving the training accuracy of 89.9\% and test set accuracy of 90.7\%. In the Monk 3 case weighted distance with just 2 non-zero coefficients gave training accuracy of 93.4\% and test result of 97.2\%.

\subsection{Hepatobiliary disorders}

The data contain four types of hepatobiliary disorders found in 536 patients of a university affiliated Tokyo-based hospital. Each case is described by 9 biochemical tests and a sex of the patient. The same 163 cases as in \cite{hayashiexp} were used as the test data. The class distribution in the training partition is 34.0\%, 23.9\%, 22.3\% and 19.8\%. This dataset has strongly overlapping classes and is rather difficult. With 49 crisp logic rules only about 63\% accuracy on the test set was achieved \cite{duch-tnn01}, and over 100 fuzzy rules based on Gaussian or triangular membership functions give about 75-76\% accuracy.

The reference \knn\ model with k=1, Euclidean distance function gave 72.7\% in the leave-one-out run on the training set (77.9\% on the test set). Although only the training set results are used in the model search results on the test set are given here to show if there is any correlation between the training and the test results. The search for the best model proceeded as follows:

{\bf First level}
\begin{enumerate}
\item Optimization of $k$ finds the best result with $k$=1, accuracy 72.7\% on training (test 77.9\%).
\item Optimization of the distance function gives training accuracy of 79.1\% with Manhattan function (test 77.9\%).
\item Selection of features removed feature Creatinine level, giving 74.3\% on the training set; (test 79.1\%).
\item Weighting of features in the Euclidean distance function gives 78.0\% on training (test 78.5\%). Final weights are [1.0, 1.0, 0.7, 1.0, 0.2, 0.3, 0.8, 0.8, 0.0].
\end{enumerate}

The best training result 79.1\% (although 77.9\% is not the best test result) is obtained by selecting the Manhattan function, therefore at the {\bf second level} this becomes the reference model:
\begin{enumerate}
\item Optimization of $k$ finds the best result with $k$=1, accuracy 72.7\% on training (test 77.9\%).
\item Selection of features did not remove anything, leaving 79.1\% on the training  (test 77.9\%).
\item Weighting of features in the Manhattan distance function gives 80.1\% on training (final weights are [1.0, 0.8, 1.0, 0.9, 0.4, 1.0, 1.0, 1.0, 1.0]; (test 80.4\%).
\end{enumerate}

At the {\bf third level} weighted Manhattan distance giving 80.1\% on training (test 80.4\%) becomes the reference model
and since optimization of $k$ nor the selection of features does not improve the training (nor test) result this becomes the final model. For comparison results of several other systems (our calculation or \cite{mitra}) on this data set are given below:

\begin{table}[htb]
\caption{Results for the hepatobiliary disorders.
Accuracy on the training and test sets.}\label{tab:hepato}
\begin{center}
\begin{tabular}{|l|r|r|}
\hline
Method 					& Training set & Test set\\
\hline
Model optimization 		& 80.1 		& 80.4 \\
FSM, Gaussian functions 	& 93 		& 75.6 \\
FSM, 60 triangular functions 	& 93 		& 75.8 \\
IB1c (instance-based) 		& -- 			& 76.7 \\
C4.5 decision tree 			& 94.4 		& 75.5 \\
Cascade Correlation 		& -- 			& 71.0 \\
MLP with RPROP 			& -- 			& 68.0 \\
Best fuzzy MLP model 		& 75.5 		& 66.3 \\
LDA (statistical) 			& 68.4 		& 65.0 \\
FOIL (inductive logic) 		& 99 		& 60.1 \\
1R (rules) 				& 58.4 		& 50.3 \\
Naive Bayes 				&-- 			& 46.6 \\
IB2-IB4 					&81.2-85.5 	& 43.6-44.6 \\
\hline
\end{tabular}
\end{center}
\end{table}

The confusion matrix is:

\be
\left(\begin{array}{rrrr}
25 &  3&  2 &  3\\
 5 & 40&  4 &  2\\
 4 &  1& 26 &  4\\
 2 &  0&  2 &  40
\end{array}\right)
\ee

Since classes strongly overlap the best one can do in such cases is to identify the cases that can be reliable classified and assign the remaining cases to pairs of classes \cite{eliminators}.

\subsection{Ionosphere data}

The ionosphere data was taken from UCI repository \cite {UCI}.
It has 200 vectors in the training set and 150 in the test set. Each data vector is
described by 34 continuous features and belongs to one of two classes.
This is a difficult dataset for many classifiers, such as decision trees: it is rather small, the number of features is quite large, the distribution of vectors among the two classes in the training set is equal, but in the test set only 18\% of the vectors are from the first class and 82\% from the second. Probably for this dataset discovery of an appropriate bias on the training set is not possible.

The reference \knn\ model with k=1, Euclidean distance function gave 86.0\% (92.0\% on the test set). The search for the best model proceeded as follows:

{\bf First level}
\begin{enumerate}
\item Optimization of $k$ finds the best result with $k$=1, accuracy 86.0\% on training (test 92.0\%).
\item Optimization of the distance function gives training accuracy of 87.5\% with Manhattan function (test 96.0\%).
\item Selection procedures leaves 10 features and gives 92.5\% on the training set; (test 92.7\%).
\item Weighting of features in the Euclidean distance function gives 94.0\% on training (test 87.3\%); only 6 non-zero weights are left.
\end{enumerate}

The best training result 94.0\% is obtained from feature weighting. Unfortunately this seems to be sufficient to overfit the data -- due to the lack of balance between the training and the test set overtraining may be quite easy. The {\bf second level} search starts from:
\begin{enumerate}
\item Optimization of $k$ does not improve the training result.
\item Optimization of the distance function gives 95.0\% with Manhattan function (test 88.0\%).
\item Selection of features did not change the training result.
\end{enumerate}

All further combinations of models reduce the training set accuracy. It is clear that there is no correlation between the results on the training and on the test set in this case.

\section{Discussion}

Meta-learning combined with the framework of similarity-based methods leads to search in the space of models derived from algorithms that are created by combining different parameters and procedures defining building blocks of learning algorithms. Although in this paper only classification problems were considered the SBM framework is also useful for  associative memory algorithms, pattern completion, missing values \cite{applied}, approximation and other computational intelligence problems.

In this paper first meta-learning results have been presented. Preliminary results with only a few extensions to the reference \knn\ model illustrated on the Monk problems, hepatobiliary disorders and the ionosphere data how the search in the model space automatically leads to more accurate solutions. Even for a quite difficult data it may be possible to find classification models that achieve 100\% accuracy on the test set. For hepatobiliary disorders a model with highest accuracy for real medical data has been found automatically. For some data sets, such as the ionosphere, there seems to be no correlation between the results on the training and on the test set.
Although the use of a validation set (or the use of the crossvalidation partitions) to guide the search process for the new models should prevent them from overfitting the data, at the same time enabling them to discover the best bias for the data other ways of model selection, such as the minimum description length (cf. \cite{Duda}), should be investigated.

Similarity Based Learner (SBL) is a software system developed in our laboratory that systematically adds various procedures belonging to the SBM framework. Methods implemented so far provide many similarity functions with different parameters, include several methods of feature selection, methods that weight attributes (based on minimization of the cost function or based on searching in the quantized weight space), methods of selection of interesting prototypes in the batch and on-line versions, and methods implementing partial-memory of the evolving system. Many optimization channels have not yet been programmed in our software, network models are still missing, but even at this preliminary stage results are very encouraging.


{\bf Acknowledgments:} Support by the Polish Committee for Scientific Research, grant no. 8 T11C 006 19, is gratefully acknowledged.


\begin{thebibliography}{99}

\bibitem{Duda}
R.O. Duda, P.E. Hart, D.G. Stork,
Pattern classification.
2nd ed, John Wiley and Sons, New York (2001)

\bibitem{statlog}
D. Michie, D.J. Spiegelhalter, C.C. Taylor,
Machine learning, neural and statistical classification.
Elis Horwood, London (1994)

\bibitem{Rohwer}
R. Rohwer, M. Morciniec,
A Theoretical and Experimental Account of n-tuple Classifier Performance.
Neural Computation {\bf 8} (1996) 657--670

\bibitem{framework}
W. Duch,
A framework for similarity-based classification methods.
In: Intelligent Information Systems VII, Malbork, Poland (1998) 288-291

\bibitem{applied}
W. Duch, R. Adamczak, G.H.F. Diercksen,
Classification, Association and Pattern Completion using Neural Similarity Based Methods.
Applied Mathematics and Computer Science {\bf 10} (2000) 101--120

\bibitem{SBMCC}
W. Duch,
Similarity-Based Methods.
Control and Cybernetics {\bf 4} (2000) xxx-yyy

\bibitem{ustron99}
W. Duch, K. Grudzi\'nski,
Weighting and selection of features in Similarity-Based Methods.
In: Intelligent Information Systems VIII, Ustro\'n, Poland (1999) 32-36

\bibitem{ijcnn99}
W. Duch, K. Grudzi\'nski,
Search and global minimization in similarity-based methods.
In: Int. Joint Conference on Neural Networks (IJCNN), Washington (1999) paper no. 742

\bibitem{RCE}
D.L. Reilly, L.N. Cooper, C. Elbaum,
A neural model for category learning.
Biological Cybernetics {\bf 45} (1982) 35--41

\bibitem{Kohonen}
T. Kohonen,
Self-organizing maps.
Springer-Verlag, Berlin Heidelberg New York (1995)

\bibitem{monks}
S.B. Thrun {\em et al.}:
The MONK's problems: a performance comparison of different learning algorithms.
Carnegie Mellon University, Technical Report  (1991) CMU-CS-91-197

\bibitem{UCI}
C.J. Mertz, P.M. Murphy,
UCI repository of machine learning datasets,\\
http://www.ics.uci.edu/AI/ML/MLDBRepository.html

\bibitem{hayashiexp}
Y. Hayashi, A. Imura, K. Yoshida,
Fuzzy neural expert system and its application to medical diagnosis.
In: 8th International Congress on Cybernetics and Systems, New York City 1990, pp. 54-61

\bibitem{duch-tnn01}
Duch W, Adamczak R, Grabczewski K,
A new methodology of extraction, optimization and application of crisp and fuzzy logical rules.
IEEE Transactions on Neural Networks 12, March 2001

\bibitem{mitra}
S. Mitra, R. De, S. Pal,
Knowledge based fuzzy MLP for classification and rule generation,
IEEE Transactions on Neural Networks 8, 1338-1350, 1997

\bibitem{eliminators}
W. Duch, R. Adamczak, Y. Hayashi,
Neural eliminators and classifiers,
7th International Conference on Neural Information Processing (ICONIP-2000), Dae-jong, Korea, Nov. 2000, ed. by Soo-Young Lee, pp. 1029 - 1034


\end{thebibliography}
\end{document}